# Herding Dynamic Weights for Partially Observed Random Field Models

**Max Welling**
Bren School of Information and Computer Science
University of California, Irvine
Irvine, USA

## Abstract

Learning the parameters of a (potentially partially observable) random field model is intractable in general. Instead of focussing on a single optimal parameter value we propose to treat parameters as dynamical quantities. We introduce an algorithm to generate complex dynamics for parameters and (both visible and hidden) state vectors. We show that under certain conditions averages computed over trajectories of the proposed dynamical system converge to averages computed over the data. Our "herding dynamics" does not require expensive operations such as exponentiation and is fully deterministic.

## 1 Introduction

It is well known that maximizing the likelihood of a Markov random field (MRF) model w.r.t. its parameters is intractable for graph structures with high treewidth. The reason is the need to compute the derivative of the normalization constant which requires inference in the MRF model. There exist a number of approximate learning techniques that attempt to address this issue (e.g. [1, 3, 15, 8, 4]). In all of these approaches the final result is a single point estimate of parameters that in some sense approximates the corresponding maximum likelihood estimate.

Bayesian inference in Markov random fields is even harder than ML estimation. To draw even a single sample from the posterior distribution one needs to compute an acceptance ratio that requires the evaluation of the partition function twice. Approximation schemes have been developed that work reasonably well for fully observed MRFs [6, 11] but they require severe approximations and it is unclear if they generalize well to MRF models with hidden variables.

In a nutshell, the following factors make traditional learning approaches "awkward": 1) At every iteration of learning the average sufficient statistics over the model need to be computed (which is highly intractable). 2) Learning is slow because it is a "double loop" algorithm with inference as the inner loop and parameter learning as the outer loop. Usually in both loops we are facing slow (linear) convergence. 3) Both inner and outer loops can can be understood as optimization problems on highly non-convex functions (the exception being fully *observed* MRFs where the likelihood surface is actually convex as a function of the parameters). 4) Once the model is learned one often needs to collect samples from it. Again, samplers can easily get stuck in local modes from which they will not mix away, leading to suboptimal final estimates of the quantities of interest. The question addressed in this paper is whether can we resolve all of these problems in one go by changing the way we view "learning".

In [10] we proposed a new perspective on "learning" fully observed MRF models that draws inspiration from findings in neuroscience, namely that synapses change their efficacy on short time scales [5]. The basic idea of this algorithm is to define a *deterministic* dynamics on the weights of a MRF model that causes the associated energy function to fluctuate in a controlled manner. In fact, at all times the local minima of this energy surface represent examples that "look like" the data that are used to drive this dynamics. In that work, the input to the algorithm was the sufficient statistics which was converted to a collection of pseudo-samples that respected those statistics and that could be used to estimate new quantities of interest. In this paper we continue this development and study random field models with hidden variables. Here the focus slightly changes because we will now be given the actual dataset (instead of just the sufficient statistics) and aim to produce hidden representations that generalize better than the original data representation.

The ideas presented in this paper are related to recent developments in learning Markov random fields using insights from stochastic approximation theory [13, 14]. The idea is that the stochastic simulation from the model (needed to compute the likelihood gradients) is never "reset". Instead, it continues where it left off before the last parameter up-date (see e.g. [7, 8, 9] for more details).

Our method departs from these learning approaches in that our goal is never to converge to a single parameter estimate. Instead, in a spirit that is somewhat akin to sampling from the posterior distribution using a Markov chain, we produce a sequence of weights and a sequence of "pseudo-samples". The latter represent low energy configurations of the fluctuating energy surface. Unlike sampling from the posterior however, the algorithm we propose is tractable. In fact, to execute herding one only needs very basic operations such as addition, multiplication and maximization. Importantly, it doesn't require exponentiation or random number generation. This could make the proposed algorithm "neurally plausible" and well suited for hardware implementation. Additional advantages include scale invariance under rescaling of the weights and the absence of "fudge factors" such as stepsize, momentum or weight decay.

## 2 Zero Temperature Limit of ML

Consider a collection of discrete random variables $(\mathbf{x}, \mathbf{z})$, where $\mathbf{x}$ will be observed and $\mathbf{z}$ will remain hidden. Subsets of variables will be denoted with $x_\alpha, z_\alpha$, where each subset is associated with a feature $g_\alpha(x_\alpha, z_\alpha)$. With each feature we also associate a weight $w_\alpha$. Given these quantities we can write the following Gibbs distribution,

$$p_w(\mathbf{x}, \mathbf{z}) = \frac{1}{Z(\mathbf{w})} \exp\left(\sum_\alpha w_\alpha g_\alpha(x_\alpha, z_\alpha)\right) \quad (1)$$

The log-likelihood $\ell$ for a dataset $\{\mathbf{x}_n\}$, $n = 1..N$ is defined as,

$$\ell = \frac{1}{N} \sum_n \log \sum_{\mathbf{z}_n} \exp(\sum_\alpha w_\alpha g_\alpha(x_{\alpha n}, z_{\alpha n})) - \log Z(\mathbf{w}). \quad (2)$$

We will denote the point estimate that maximizes this log-likelihood with $\mathbf{w}^*$.

Introducing a variational posterior distribution $Q_n(\mathbf{z}_n)$ we rewrite $\ell$ as,

$$\ell = \frac{1}{N} \sum_n \max_{Q_n} \left[\sum_\alpha w_\alpha \mathbb{E}[g_\alpha(x_{\alpha n}, z_{\alpha n})]_{Q_n} + \mathcal{H}(Q_n)\right] \\ - \log Z(\mathbf{w}) \quad (3)$$

where $\mathcal{H}(P) = -\sum_\mathbf{y} P(\mathbf{y}) \log P(\mathbf{y})$ is the entropy of $P$.

In a similar spirit, the second term can also be written variationally by introducing distributions $R(\mathbf{z}, \mathbf{x})$,

$$\ell = \frac{1}{N} \sum_n \max_{Q_n} \left[\sum_\alpha w_\alpha \mathbb{E}[g_\alpha(x_{\alpha n}, z_{\alpha n})]_{Q_n} + \mathcal{H}(Q_n)\right] \\ - \max_R \left[\sum_\alpha w_\alpha \mathbb{E}[g_\alpha(x_\alpha, z_\alpha)]_R + \mathcal{H}(R)\right] \quad (4)$$

We can define a joint function $\tilde{\ell}(\{Q_n\}, R, \mathbf{w})$ by removing the maximization operations in the expression above. It's relation (for any $\{Q_m, R\}$) to $\ell(\mathbf{w})$ is then given by

$$\tilde{\ell} = \ell - \frac{1}{N} \sum_n KL[Q_n(\mathbf{z}_n) || P_w(\mathbf{z}_n|\mathbf{x}_n)] \\ + KL[R(\mathbf{x}, \mathbf{z}) || P_w(\mathbf{x}, \mathbf{z})] \quad (5)$$

We can interpret the variational expression of Eqn.4 as a maximum entropy problem with hidden variables. In particular, we can reorder to find,

$$\ell = \max_{\{Q_n\}} \min_R \ \frac{1}{N} \sum_n \mathcal{H}(Q_n) - \mathcal{H}(R) \\ + \sum_\alpha w_\alpha \left(\frac{1}{N} \sum_n \mathbb{E}[g_\alpha(x_{\alpha n}, z_{\alpha n})]_{Q_n} - \mathbb{E}[g_\alpha(x_\alpha, z_\alpha)]_R\right) \quad (6)$$

In this expression the weights $w_\alpha$ act as Lagrange multipliers enforcing the constraint: $\frac{1}{N} \sum_n \mathbb{E}[g_\alpha(x_{\alpha n}, z_{\alpha n})]_{Q_n} = \mathbb{E}[g_\alpha(x_\alpha, z_\alpha)]_R$. Note that in addition to satisfying these constraints this optimization problem seeks to achieve high entropy for the distributions $\{Q_n\}$ and $R$. However the entropy for $\{Q_n\}$ and $R$ have reversed signs in the objective leading to a minimax problem.

We now introduce a temperature by replacing $w_\alpha \to w_\alpha/T, \forall \alpha$. Taking the limit $T \to 0$ of $\ell_T \triangleq T\ell$ we see that the entropy terms vanish. For a given value of $\mathbf{w}$ and in the absence of entropy, the optimal distributions $\{Q_n\}$ and $R$ are delta-peaks and their averages can thus be replaced with maximizations, resulting in the objective,

$$\ell_0(\mathbf{w}) = \frac{1}{N} \sum_n \max_{\mathbf{z}_n} \left[\sum_\alpha w_\alpha g_\alpha(x_{\alpha n}, z_{\alpha n})\right] \\ - \max_\mathbf{s} \sum_\alpha [w_\alpha g_\alpha(s_\alpha)] \quad (7)$$

where we renamed $s_\alpha = (x_\alpha, z_\alpha)$.

In the next section we will study this function more closely.

## 3 Tipi Functions

The results derived in this section are in close analogy to the results obtained in [10] but also have important differences.

First we define $\{\mathbf{z}_n^*, \mathbf{s}^*\}$ to be the values that maximize their respective terms in Eqn.7. One can then easily observe that $\ell_0$ is locally linear in $\mathbf{w}$ and that its derivative w.r.t. $w_\alpha$ is given by,

$$\nabla_{w_\alpha} \ell_0 = \left[\frac{1}{N} \sum_n g_\alpha(x_{\alpha n}, z_{\alpha n}^*)\right] - g_\alpha(s_\alpha^*) \quad (8)$$

Note that $\sum_\alpha w_\alpha \nabla_{w_\alpha} \ell_0 = \ell_0$.

The following properties hold for $\ell_0$.

**P1.** $\ell_0$ *is continuous piecewise linear ($C^0$ but not $C^1$)*. It is clearly linear in $\mathbf{w}$ as long as the states $\{\mathbf{z}_n^*, \mathbf{s}^*\}$ that minimize their respective energies don't change. However, for different values of $\mathbf{w}$ we may have different "maximizing" states $\{\mathbf{z}_n^*, \mathbf{s}^*\}$ implying that the derivative 8 changes discontinuously. Note however that $\ell_0$ itself is continuous because it is the difference between two maximizations over a set hyper-planes.

**P2.** $\ell_0$ *is a concave, non-positive function of $\mathbf{w}$ with a maximum at $\ell_0(0) = 0$*. This is true because we maximize over all variables in the second term, while in the first we clamp data-points to $\mathbf{x}$. Therefore, $\ell_0 \leq 0$. If we furthermore assume that for any direction in weight space there are always two data-cases with different energy, we will have $\ell_0 < 0$ outside the origin which means that the maximum at $\mathbf{w} = 0$ is unique (see also section 7). Concavity can be proven by using a convex combination of weights $\sum_j p_j \mathbf{w}_j$ and pulling the summations outside the "max" operations. Again, because the second term maximizes over more variables than the first term, the total result will be a decrease in $\ell_0$ which proves concavity.

**P3.** $\ell_0$ *is scale free*. This follows because $\ell_0(\beta \mathbf{w}) = \beta \ell_0(\mathbf{w})$ as can be easily checked. This means that the function has exactly the same structure at any scale of $\mathbf{w}$.

The above properties justify the name "Tipi" function[1] (see Figure 1).

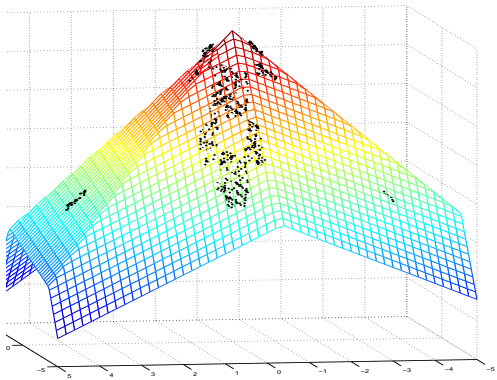

Figure 1: Two-dimensional Tipi-function for herding with features $f(x) = sin(x)$ and $g(x) = cos(x)$, $x = [-\pi, \pi + 1, .., 2.86]$ and a uniform distribution $P(x)$ to compute $\bar{f}$ and $\bar{g}$. Small dots represent weights sampled during herding.

In the next section we will define a dynamical system, which we call "herding", as gradient ascent on a Tipi function with unit (or rather arbitrary) stepsize.

---

[1] A Tipi is a native Indian dwelling.

## 4 Herding

We first note that due to the zero temperature limit it has become pointless to try to locate the maximum of $\ell_0$ (it is located at $\mathbf{w}^* = 0$ as argued under P2). This is in stark contrast to the original ML problem where we were seeking to maximize $\ell_1$. Instead we propose to run a gradient ascent algorithm on $\ell_0$ with a fixed stepsize (the actual value of which will turn out not to matter). Keeping Figure 1 in mind, the gradient is locally constant (and not defined on the boundaries between flat faces). Due to the fact that $\sum_\alpha w_\alpha \nabla_{w_\alpha} \ell_0 = \ell_0 < 0$ we see that the gradient has a component pointing towards the origin. In taking a step along the local gradient we may enter a new linear face and $\ell_0$ may decrease as a result. So, due to the *finite stepsize* $\eta$ one will never converge to $\mathbf{w} = 0$. In the following we will set $\eta = 1$ because, as we will prove in section 8, the sequence of states $\mathbf{s}_t, t \geq 0$ does not depend it.

The herding equations are thus,

$$\mathbf{z}_{nt}^* = \arg\max_{\mathbf{z}_n} \sum_\alpha w_{\alpha,t-1} \, g_\alpha(x_{\alpha n}, z_{\alpha n}) \; \forall \, n \qquad (9)$$

$$\mathbf{s}_t^* = \arg\max_{\mathbf{s}} \sum_\alpha w_{\alpha,t-1} \, g_\alpha(s_\alpha) \qquad (10)$$

$$w_{\alpha t} = w_{\alpha,t-1} + \left[\frac{1}{N}\sum_{n=1}^N g_\alpha(x_{\alpha n}, z_{\alpha nt}^*)\right] - g_\alpha(s_{\alpha t}^*) \qquad (11)$$

These equations are similar to herding for the fully observed case [10], but different in the sense that we need to impute the unobserved variables $\mathbf{z}_n$ for every data-case separately through maximization. The weight update also consist of a positive "driving term", which is now a changing average over data-cases, and a negative term which is identical to the corresponding term in the fully observed case.

We like to emphasize that the dynamical system is entirely deterministic. This has advantages in terms of computation, since random number generation can be expensive. The equations thus produce pseudo-samples that look random, but should not be interpreted as random samples. Also, the weights generated during the execution of the algorithm should not be interpreted as samples from some Bayesian posterior distribution.

When executing the herding equations one can empirically observe that irrespective of the initial condition every weight configuration will converge on a unique global attractor set (or invariant set). This means that, in close analogy to Markov chains, the dynamical system will forget its initial condition. Another empirical observation is that the dynamical system mixes very rapidly over the attractor set which gives it a key advantage over sampling using a Markov chain. There are many questions that remain to be studied in relation to this attractor set. Is it unique? Does it have fractal dimension? Can one define an invariant measure? Is this measure unique and/or ergodic?

Interestingly, the tools to investigate these questions are to be found in a field of mathematics not usually associated with machine learning, namely that of complex dynamical system and chaos theory. In fact, the herding equations are similar to dynamical systems studied in the mathematical literature known as "piecewise isometries" [2]. Some interesting properties have been proven about these systems such as the possibility of fractal attractor sets in weight space (so called strange attractors), the fact that the orbits usually have infinite period, the fact that all Lyapunov exponents are 0 and the fact that the number of realizable sequences of length $n$, i.e. $[\mathbf{s}_{t+1}, ..., \mathbf{s}_{t+n}]$, grows polynomially with $n$. These type of models are known in the physics literature as "critical", or "edge of chaos" and have some intriguing relations to the dynamics of firing neurons, earthquakes, sand-piles and forest fires.

In the next section we prove an important property of herding which will clarify in what sense the sequence of weights represents a model of the data.

## 5 Ergodicity

Why do we think that the herding algorithm from the previous section produces anything useful? To partially answer that question we will now turn our attention to the properties of the sequence of states $\mathbf{s}_t$, $t = 1..\infty$. We claim that under some mild conditions, time averages over these states converge to ensemble averages over the data as expressed by the following proposition:

**Proposition 1:** If $\forall \alpha \ \lim_{\tau \to \infty} \frac{1}{\tau} w_{\alpha\tau} = 0$, then

$$\lim_{T \to \infty} \frac{1}{T} \sum_{t=1}^{T} g_\alpha(s_{\alpha t}) \to \lim_{T \to \infty} \frac{1}{T} \sum_{t=1}^{T} \bar{g}_{\alpha t} \quad (12)$$

with $\bar{g}_{\alpha t} = \frac{1}{N} \sum_{n=1}^{N} g_\alpha(x_{\alpha n}, z^*_{\alpha n t})$.

**Proof:**
$$\delta w_{\alpha t} = \bar{g}_{\alpha t} - g_\alpha(s_{\alpha t}) \quad (13)$$

with $\delta w_{\alpha t} \triangleq w_{\alpha t} - w_{\alpha, t-1}$. Next, average left and right hand sides over t,

$$\frac{1}{\tau} \sum_{t=1}^{\tau} \delta w_{\alpha t} = \frac{1}{\tau}(w_{\alpha\tau} - w_{\alpha 0}) = \frac{1}{T} \sum_{t=1}^{T} \bar{g}_{\alpha t} - \frac{1}{T} \sum_{t=1}^{T} g_\alpha(s_{\alpha t}) \quad (14)$$

Using the premise that the weights grow slower than linearly we see that the left hand term vanishes in the limit $\tau \to \infty$ which proves the result.

The premise that the weights grow slower than linear does *not* mean that we have to be able to solve the difficult optimization problems of Eqn.9 and 10. Partial progress in the form of a few iterations of coordinate-wise ascent is often enough to keep the weights finite. In section 7 we define a fully tractable version of herding for which the premise to proposition 1 holds.

These consistency equations are in direct analogy to the maximum likelihood (ML) equations of Eqn.1 for which the following moment matching conditions hold at the ML estimate $w^*$ and for all $\alpha$,

$$\frac{1}{N} \sum_{n=1}^{N} \mathbb{E}[g_\alpha(x_{\alpha n}, z_{\alpha n})]_{p_w(z_n | x_n)} = \mathbb{E}[g_\alpha(x_\alpha, z_\alpha)]_{p_w(z, x)} \quad (15)$$

These consistency conditions alone are not sufficient to guarantee a good model. After all, the dynamics could simply ignore the hidden variables by keeping them constant and still satisfy the matching conditions. In this case the hidden and visible subspaces completely decouple defeating the purpose of using hidden variables in the first place. Note that the same holds for the ML consistency conditions. However, a ML solution also strives for high entropy in the hidden states. We believe that the herding dynamics similarly induces entropy in the distributions for $\mathbf{z}$ avoiding the decoupling phenomenon described above.

## 6 Recurrence

The premise for proposition 1 is that the weights do not run away linearly. Therefore, if we can show that the weights are contained in a compact set around the origin, then the premise certainly holds. In this section we will assume that we can find the global maximum for both the $\{z_n\}$ variables as well as for the $\mathbf{s}$ variables. (We call this algorithm "idealized herding" in the following.) This assumption is unrealistic for many real problems but in the next section we will show that we can relax this condition by formulating a fully tractable version of herding for which the premise will also hold. Many experiments were conducted by using a local optimization algorithm initialized at the state from the previous time step. In almost all cases, this was sufficient to keep the weights small (but lacks the guarantees). The intuitive reason is that the energy for states that are initially hard to reach with local search, will keep on growing until the local algorithm will be able to reach it. This same argument was used in [8] to motivate his learning algorithm based on stochastic approximation.

We will now show that the weights in idealized herding stay contained inside a compact set. We recall property P2 from section 3, $\sum_\alpha w_\alpha \nabla_{w_\alpha} \ell_0 = \ell_0 < 0$. We first prove a lemma stating that the norm of the gradient of $\ell_0$ is bounded from above.

**Lemma 1:** If $|g_\alpha(s_\alpha)| < \infty$, $\forall \mathbf{s}, \alpha$, then $\exists \mathcal{B}$ such that $||\nabla \ell_0||_2 < \mathcal{B}$.

**Proof:** $\nabla_{w_\alpha} \ell_0(\mathbf{w}) = \frac{1}{N} \sum_{n=1}^{N} g_\alpha(z^*_{\alpha n}, x_{\alpha n}) - g_\alpha(s^*_\alpha)$

with $\{\mathbf{z}_n^*, \mathbf{s}^*\}$ the maximizing states. Since all $g_\alpha(\mathbf{s}_\alpha)$ are finite for any value of $\mathbf{s}, \alpha$ the norm of the gradient must be bounded as well.

We now prove that there will be some radius $\mathcal{R}$ such that the herding algorithm will always decrease the norm $||\mathbf{w}||_2$.

**Proposition 2:** $\exists$ radius $\mathcal{R}$ such that an idealized herding update performed outside this radius, will always decrease the norm $||\mathbf{w}||_2$.

**Proof:** Write the herding update as $w'_\alpha = w_\alpha + \nabla_{w_\alpha}\ell_0$. Take the inner product with $w'_\alpha$ leading to, $||\mathbf{w}'||_2^2 = ||\mathbf{w}||_2^2 + 2\sum_\alpha w_\alpha \nabla_{w_\alpha}\ell_0 + ||\nabla_{w_\alpha}\ell_0||_2^2$, which leads to $\delta||\mathbf{w}||_2^2 < 2\ell_0 + \mathcal{B}^2$. We now use the fact that 1) $\ell_0 < 0$ outside the origin (P2), 2) $\mathcal{B}$ is constant (i.e. doesn't scale with $\mathbf{w}$) and 3) the scaling property $\ell_0(\beta\mathbf{w}) = \beta\ell_0(\mathbf{w})$ (P3) to argue that there is always some radius $\mathcal{R}$ for which $\delta||\mathbf{w}||_2 < 0, \forall ||\mathbf{w}||_2 > \mathcal{R}$ (if not, increase $\beta$ by a sufficient amount).

**Corollary:** $\exists$ radius $\mathcal{R}'$ such that a herding algorithm initialized inside a ball with radius $\mathcal{R}'$ will never generate weights $\mathbf{w}$ with norm $||\mathbf{w}||_2 > \mathcal{R}'$.

This follows because in the worst case we could still take one step radially outward starting somewhere on the surface of the ball at radius $\mathcal{R}$. Since the gradient is bounded in magnitude by $\mathcal{B}$ we have that $\mathcal{R}' \leqq \mathcal{R} + \mathcal{B}$.

## 7 A Tractable Version of Herding

The results of the previous section were only valid for the case where we could find the global minimum of the energy function. This begs the question: "did we not replace one intractable problem for another?" The claim we have been making throughout this paper is that much simpler (tractable) optimization schemes, such as local minimizers are usually sufficient to contain the weights into a compact region. We reemphasize that different optimizations schemes will lead to different invariant attractor sets and hence possibly different generalization behavior. This in spite of the fact that the constraints on the training data are recovered on average for all herding variants that do not lead to run-away dynamics of the weights.

In this section we will propose a fully tractable herding variant that is guaranteed to remain in a compact region of weight space under very mild conditions.

**Proposition 3:** Call $\mathcal{A}$ any tractable optimization algorithm to locate a local minimum in the energy function. This algorithm will be used to compute both $z_n^*$ and $s^*$. Call $\mathcal{E}_\mathcal{A}(x_n, \mathbf{w}) = -\sum_\alpha w_\alpha g_\alpha(x_n, z_n^*)$ the energy of data-case $n$ (note that this definition depends on the algorithm $\mathcal{A}$). Assume that for every direction in weight space there are at least two data-vectors that have a different energy. Then the following tractable herding algorithm will remain in a compact region of weight space: Apply the usual herding updates with the difference that the optimization for $\mathbf{s}$ is initialized at the state $(x_{n^*}, z_{n^*}^*)$ which represents the data-case with lowest energy $\mathcal{E}_\mathcal{A}(x_{n^*}, z_{n^*}^*)$.

**Proof:** If for every direction in weight space there are least two data-cases with a different energy, then the average over those energies will be higher than their minimum. Since energies simply scale by changing the norm of the weight vector the result thus holds for all weight vectors. We now use that $\sum_\alpha w_\alpha \nabla_{w_\alpha}\ell_0(\mathbf{w}) = \ell_0(\mathbf{w})$, and the fact that $\ell_0(\mathbf{w})$ precisely equals the energy gap, to argue that for every $\mathbf{w}$ the gradient has a negative inner product with the weight vector itself. Call $\hat{\mathbf{w}} = \mathbf{w}/||\mathbf{w}||$ the normalized weight vector and $\ell_0(\hat{\mathbf{w}}^*)$ the minimal energy gap over all normalized weight vectors $\hat{\mathbf{w}}$. By scaling the weight-vector $\mathbf{w} = \beta\hat{\mathbf{w}}$ the minimal energy-gap will also scale: $\ell_0(\beta\hat{\mathbf{w}}^*) = \beta\ell_0(\hat{\mathbf{w}}^*)$ and can thus be made arbitrarily negative. Now using the fact that the gradients are bounded in norm and using the same arguments as in the proof of proposition 2, the result follows.

Note that the premise for this algorithm is the same as that for herding with full maximization. Since $\ell_0$ has the form of a Tipi function any projection for which all data-cases have the same energy would lead to a flat (horizontal) face in the Tipi function. On this face the gradients would vanish and any form of herding would simply stop there. One could encounter this degenerate Tipi function when using trivial (e.g. constant) features. To recover from such an issue one should simply prune these features away.

The algorithm described in this section requires one to query all data-vectors. This makes sense in the case of herding with hidden variables because then our objective is to produce superior hidden representations of the original data and the algorithm requires access to all data-cases anyway. Ironically, for the simpler fully observed problem [10] this approach doesn't apply because there we only assumed having access to the sufficient statistics of the data and not the full dataset itself.

## 8 Invariant Transformations

We have already argued that the only effect of changing the stepsize $\eta$ is to change the scale of the invariant attractor set of the sequence $w_{\alpha t}$. More precisely, denote $v_{\alpha t}, t > 0$ the standard herding sequence with stepsize $\eta = 1$ and $w_{\alpha t}, t > 0$ the sequence with an arbitrary stepsize $\eta$. We then claim that if we initialize $v_{\alpha,t=0} = \frac{1}{\eta}w_{\alpha,t=0}$ and apply the respective herding updates for $w_\alpha$ and $v_\alpha$ afterwards, the relation $v_{\alpha t} = \frac{1}{\eta}w_{\alpha t}$ will remain true for all $t > 0$ and in particular, the states $\mathbf{s}_t$ will be the same for both sequences (for proof see below).

More generally, we will call two attractor sets equivalent if we can find weight initializations for their respective herd-ing updates such that the state sequences are equal. Of course, if such a transformation does exist but one initializes both sequences with arbitrary different values, then the state sequences will not be identical. However, if one accepts the conjecture that there is a unique invariant attractor set, then this difference can be interpreted as a difference in initialization which only affects the transient behavior (or "burn-in" behavior) but not the (marginal) distribution $p(\mathbf{s})$ from which the states $\mathbf{s}_t$ will be sampled.

One can now ask the question: "*what group of transformations on the herding equations will result in equivalent attractor sets*". In the proposition below we define a set of transformations that satisfies equivalence in the above sense.

**Proposition 4:** The following parameterized family of herding equations are equivalent[2]:

$$w_{\alpha t} = w_{\alpha,t-1} + \eta \left( \frac{1}{N} \sum_{n=1}^{N} g_\alpha(x_n, z^*_{n,t-1}) - g_\alpha(s^*_{t-1}) \right) \tag{16}$$

$$z^*_{nt} = \arg\max_{z_n} \sum_\alpha [\gamma(w_{\alpha t} + a_\alpha)] g_\alpha(x_n, z_n) \tag{17}$$

$$s^*_t = \arg\max_s \sum_\alpha [\gamma(w_{\alpha t} + a_\alpha)] g_\alpha(s) \tag{18}$$

**Proof:** Define the new variables $v_\alpha = \frac{1}{\eta}(w_\alpha + a_\alpha)$. In terms of these the above updates become,

$$\eta v_{\alpha t} - a_\alpha = \tag{19}$$
$$\eta v_{\alpha,t-1} - a_\alpha + \eta \left( \frac{1}{N} \sum_{n=1}^{N} g_\alpha(x_n, z^*_{nt}) - g_\alpha(s^*_{t-1}) \right)$$

$$z^*_{nt} = \arg\max_{z_n} \sum_\alpha \gamma \eta v_{\alpha t}\, g_\alpha(x_n, z_n) \tag{20}$$

$$s^*_t = \arg\max_s \sum_\alpha \gamma \eta v_{\alpha t}\, g_\alpha(s) \tag{21}$$

which, after canceling terms and removing unnecessary factors inside maximizations become equivalent to the standard herding equations with $\eta' = \gamma' = 1, a'_\alpha = 0$.

Note that we are not allowed to use $\alpha$-dependent parameters $\eta_\alpha$ or $\gamma_\alpha$ because we can no longer pull them out of the maximizations in steps 20 and 21. This observation implies that changing the *relative* stepsizes between the updates has an effect on the invariant attractor set and thus potentially on the generalization performance. On the other hand, changing the overall stepsize, or the overall scale of the energy (i.e. the temperature) or even adding a constant (but $\alpha$ dependent) offset to the weights will not change the generalization performance.

---

[2]Note that we slightly simplified notation by removing the cluster indices from the states and omitting $\forall \alpha$ etc.

The above analysis has important implications for any attempts to combine herding of "fast weights" with learning of "slow weights" [9]. *Adding* slow and fast weights seems doomed in the context of herding because herding will simply compensate by shifting the attractor set. In fact, it suggests *multiplying* slow and fast weights because that will translate into a different and perhaps better herding sequence[3].

## 9 Learning Versus Herding

The herding algorithm discussed in the previous section simply turns data into an unlimited number of pseudo-samples (denoted with $\{s_i\}$ above). It can thus be thought of as a kind of filter. Can we use these samples to retain information for the long term? The situation is reverse to what we are used to in machine learning where we first learn the weights and subsequently sample from this model to make inferences.

First note that we need the data to drive herding with hidden variables. A "model" is defined to be a herding algorithm that operates without direct access to the data. Instead it uses a collection of summary statistics. Data enters herding in the positive term of Eqn.11, so to run herding without data we need to estimate the *rate* with which to drive the weights. This can be achieved through the following online averaging process:

$$r_{\alpha t} = \frac{t-1}{t} r_{\alpha,t-1} + \frac{1}{t} \bar{g}_{\alpha t}\ , t = 1,.. \tag{22}$$

with $\bar{g}_{\alpha t} = \frac{1}{N} \sum_{n=1}^{N} \sum_\alpha w_{\alpha t} g_\alpha(x_{\alpha n}, z^*_{\alpha nt})$ and $r_{\alpha 0} = 0$. Once the learning phase has finished, we can decouple the data and run herding with the rate functions $r_\alpha$ instead. Note however that this is an approximation since the exact positive term is an implicit function of the weights through $z^*_n(\mathbf{w})$ which we now replace with a constant function. One could imagine learning more flexible regression functions $r_\alpha(\mathbf{w})$ to approximate $\bar{g}_\alpha(\mathbf{w})$.

Examples of constant "rate filters", which drive the weights connecting visible units and hidden units are given in Figure 2 for the restricted Boltzman machine described in section 10. These filters represent transformations, that turn one digit into another. They visualize the correlations modeled by the hidden units.

This new type of "learning with dynamic synapses" may provide an exciting alternative view of learning in the brain. Indeed, some results in the neuroscience literature point to the fact that synaptic efficacies change on short timescales [5]. The computational framework of herding is one possible working model of how to compute with these dynamic synapses.

---

[3]In [9] the slow and fast weights are added, but the fast weights are not exactly driven by herding.

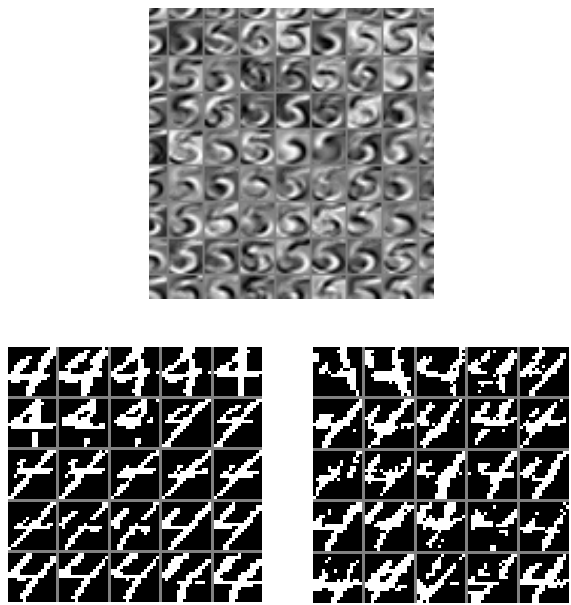

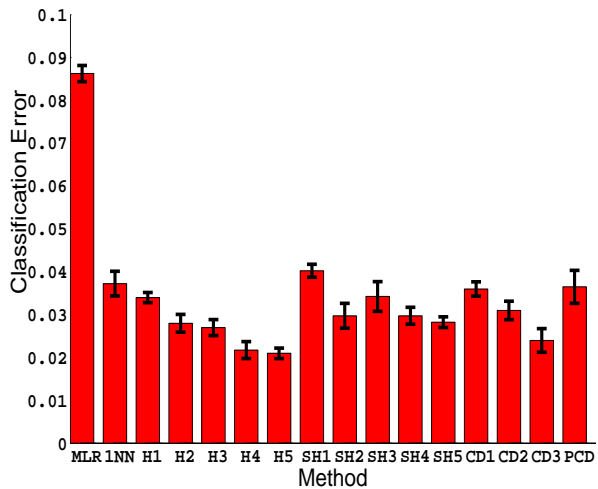

Figure 2: TOP: "Rate-filters" for digit "5" (Eqn.22 between a hidden variable and all observed variables in a restricted Boltzman machine (see section 10). Note that these are *not* average weights $W_{ij}$ (which would look like random noise). Rate-filters correspond to the average correlations $s_i s_j$ between hidden and visible states estimated from the pseudo-samples. BOTTOM LEFT: A sequence of (visible) states recorded during herding for the digit "4". Note the smooth transitions between confabulated digits. BOTTOM RIGHT: A similar herding sequence for the digit "4" but now using learned "rate functions". Decoupling the data from herding clearly makes the fantasized digits more "noisy".

## 10 Experiments

We studied herding on the architecture of a restricted Boltzman machine [3]. We used features $g_\alpha(z_\alpha, y_\alpha) = \{z_i, y_j, z_i y_j\}$ and the $\{-1, +1\}$ representation because we found it worked significantly better than the $\{0, 1\}$ representation. To increase the entropy of the hidden units we left out the growth update for the features $\{z_i\}$ implying that $p(z_i = 1) \approx 0.5$. The intuition is the same as for bagging: we want to create a high diversity of (almost independent) ways to reconstruct the data because it will reduce the variance when making predictions. We observed that high entropy hidden representations automatically emerged when using a large number of hidden units. In contrast, for a small number of hidden units (say $K < 30$) there is a tendency for the system to converge on low entropy representations and the trick delivers some improvement.

We applied herding to the USPS Handwritten Digits dataset[4] which consists of 1100 examples of each digit 0 through 9 (totalling 11,000 examples). Each image has 256 pixels and each pixel has a value between [1..256] which we turned into a binary representation through the mapping

---

[4] Downloaded from *http://www.cs.toronto.edu/~roweis/data.html*

Figure 3: Classification results on USPS digits. 700 digits per class were used for training, 300 for validation and 100 for testing. Shown are average results over 4 different splits and their standard errors. From left to right: MLR (multinomial logistic regression), 1NN (1-nearest neighbor), H1-H5 (herding using local optimization with 50,100,250,500 and 1000 hidden units respectively), SH1-SH5 (safe, tractable herding from section 7 with 50,100,250,500 and 1000 hidden units respectively), CD1-CD3 (contrastive divergence with 50,100,250 hidden units respectively) and PCD (persistent CD with 500 hidden units).

$X'_i = 2\Theta(X_i/256 - 0.2) - 1$ with $\Theta(x > 0) = 1$ and 0 otherwise. Each digit class was randomly split into 700 train, 300 validation and 100 test examples. As benchmarks we used 1NN using Manhattan distance and multinomial logistic regression, both in pixel space.

We used two versions of herding, one where the maximization over **s** was initialized at the value from the previous time step (H) and one where we initialize at the data-case with the lowest energy (SH – the algorithm from section 7). In both cases we ran herding for 2000 iterations for each class individually. During the second 1000 iterations we computed the energies for the training data in that class, as well as for all validation and test data across all classes. At each iteration we then used the training energies to standardize the validation and test energies by computing their Z-scores: $E'_i = (E_i - \mu_{\text{trn}})/\sigma_{\text{trn}}$ where $\mu_{\text{trn}}$ and $\sigma_{\text{trn}}$ represent the mean and standard deviation of the energies of the training data at that iteration. The standardized energies for test and validation data were subsequently averaged over herding iterations (using online averaging). Once we have collected these average standardized energies across all digit classes we fit a multinomial logistic regression classifier to the validation data, using the 10 class-specific energies as features.

We also compared these results against models learned with contrastive divergence [3] (CD) and persistent CD [8]

(PCD). For both CD and PCD we first applied (P)CD learning for 1000 iterations in batch mode, using a stepsize of $\eta = 1E - 3$. A momentum parameter of 0.9 and 1-step reconstructions were used for CD. No momentum and a single sample in the negative phase was used for PCD. In the second 1000 iterations we continued learning but also collected standardized validation and test energies as before which we subsequently used for classification. We have also experimented with chains of length 10 and found that it didn't improved the results but became prohibitively inefficient. To improve efficiency we experimented with learning in mini-batches but this degraded the results significantly, presumably because the number of training examples used to standardize the energy scores became less reliable.

The results reported in Figure 3 show the classification results averaged across 4 runs with different splits and for different values of hidden units. Without trying to claim superior performance we merely want to make the case that herding can be leveraged to achieve state-of-the-art performance (note that USPS error rates are higher than MNIST error rates). We also see that the tractable version of herding did not perform as well as the herding using local optimization, which in turn performed equally well as learning a model using CD. Persistent CD did not give very good results presumably because we did not use optimal settings for step-size, weight-decay etc. It is finally interesting to observe that there does not seem to be any sign of overfitting for herding. For the model with 1000 hidden units, the total number of *real* parameters involved is around 1.5 million which represents more capacity than the 1.5 million *binary* pixel values in the data.

## 11 Discussion

Herding reminds us of a form of unsupervised bagging or boosting where many weak models are combined into one powerful predictor (see also [12]). It does not rely on expensive operations such as random number generation and exponentiation and is therefore ideally suited for hardware implementation. Finally, there may be exciting connections to the concept of dynamical weights in neuroscience.

Herding effectively considers weights as a kind of nuisance parameters. It can be seen as a procedure to "marginalize out" these weights resulting in a non-parametric encoding of the model, namely as a collection of samples. Deep learning is a suitable playing field for herding since for this task one cares more about the representations of the hidden layers than about the weights. Stacking RBMs seems straightforward in theory but many hurdles may have to be overcome before "deep herding" will become practical. Another obvious direction for future research is herding in the context of conditional random fields.

## Acknowledgements

This material is based upon work supported in part by the National Science Foundation under Award Number IIS-0447903 and IIS-0535278 and by ONR-MURI under Grant No. 00014-06-1-073. We thank Y. Bengio, R. Palais, R. Thibaux and A. Gorodetski for discussion.

## References

[1] J. Besag. Efficiency of pseudo-likelihood estimation for simple Gaussian fields. *Biometrika*, 64:616–618, 1977.

[2] A. Goetz. Dynamics of piecewise isometries. *Illinois Journal of Math*, 44:3:465–478, 2000.

[3] G.E. Hinton. Training products of experts by minimizing contrastive divergence. *Neural Computation*, 14:1771–1800, 2002.

[4] A. Hyvarinen. Estimation of non-normalized statistical models using score matching. *Journal of Machine Learning Research*, 6:695–709, 2005.

[5] W. Maass and A. M. Zador. Dynamic stochastic synapses as computational units. In *Advances in Neural Information Processing Systems*, pages 903–917. MIT Press, 1998.

[6] I. Murray and Z. Ghahramani. Bayesian learning in undirected graphical models: approximate MCMC algorithms. In *Proceedings of the 14th Annual Conference on Uncertainty in AI*, pages 392–399, 2004.

[7] R.M. Neal. Connectionist learning of belief networks. *Articial Intelligence*, 56:71–113, 1992.

[8] T. Tieleman. Training restricted boltzmann machines using approximations to the likelihood gradient. In *Proceedings of the International Conference on Machine Learning*, volume 25, pages 1064–1071, 2008.

[9] T. Tieleman and G.E. Hinton. Using Fast Weights to Improve Persistent Contrastive Divergence. In *Proceedings of the International Conference on Machine Learning*, volume 26, 2009.

[10] M. Welling. Herding dynamical weights to learn. In *Proceedings of the 21st International Conference on Machine Learning*, Montreal, Quebec, CAN, 2009.

[11] M. Welling and S. Parise. Bayesian random fields: The Bethe-Laplace approximation. In *Proc. of the Conf. on Uncertainty in Artificial Intelligence*, pages 512–519, 2006.

[12] M. Welling, R. Zemel, and G.E. Hinton. Self-supervised boosting. In *Neural Information Processing Systems*, volume 15, Vancouver, Canada, 2002.

[13] L. Younes. Parametric inference for imperfectly observed gibbsian fields. *Probability Theory and Related Fields*, 82:625–645, 1989.

[14] A.L. Yuille. The convergence of contrastive divergences. In *Advances in Neural Information Processing Systems*, volume 17, pages 1593–1600, 2004.

[15] S.C. Zhu and X.W. Liu. Learning in Gibbsian fields: How accurate and how fast can it be? *IEEE Trans. on Pattern Analysis and Machine Intelligence*, 24(7):1001–1006, 2002.